# TRANSPARENCY ATTACKS: HOW IMPERCEPTIBLE IMAGE LAYERS CAN FOOL AI PERCEPTION

Forrest McKee[1] and David Noever[2]
PeopleTec, 4901-D Corporate Drive, Huntsville, AL, USA, 35805
[1]forrest.mckee@peopletec.com       [2] david.noever@peopletec.com

***ABSTRACT***

*This paper investigates a novel algorithmic vulnerability when imperceptible image layers confound multiple vision models into arbitrary label assignments and captions. We explore image preprocessing methods to introduce stealth transparency, which triggers AI misinterpretation of what the human eye perceives. The research compiles a broad attack surface to investigate the consequences ranging from traditional watermarking, steganography, and background-foreground miscues. We demonstrate dataset poisoning using the attack to mislabel a collection of grayscale landscapes and logos using either a single attack layer or randomly selected poisoning classes. For example, a military tank to the human eye is a mislabeled bridge to object classifiers based on convolutional networks (YOLO, etc.) and vision transformers (ViT, GPT-Vision, etc.). A notable attack limitation stems from its dependency on the background (hidden) layer in grayscale as a rough match to the transparent foreground image that the human eye perceives. This dependency limits the practical success rate without manual tuning and exposes the hidden layers when placed on the opposite display theme (e.g., light background, light transparent foreground visible, works best against a light theme image viewer or browser). The stealth transparency confounds established vision systems, including evading facial recognition and surveillance, digital watermarking, content filtering, dataset curating, automotive and drone autonomy, forensic evidence tampering, and retail product misclassifying. This method stands in contrast to traditional adversarial attacks that typically focus on modifying pixel values in ways that are either slightly perceptible or entirely imperceptible for both humans and machines.*

***KEYWORDS***

*Adversarial attacks, computer vision, transformers, GPT4, object detection*

## 1. INTRODUCTION

The exploration of adversarial vulnerabilities in computer vision systems has unveiled a novel class of attacks that exploit the dichotomy between human and machine perception, mainly by manipulating imperceptible transparency layers or alpha channels of the Portable Network Graphic (PNG), a type of raster image file cast in red, green, blue, alpha (RGBA) channels (**Figure 1**). In our case, the RGB channels hold a grayscale background, and the alpha channel holds the white-blended grayscale attack foreground, which is semi-transparent and either not examined or flattened by a broad range of vision algorithms.

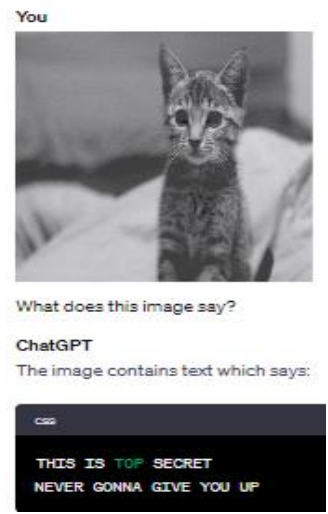

*Figure 1. Transparency attack on Vision-GPT4*

The uniqueness of this approach lies in its ability to manipulate images without altering their apparent visual content, thus presenting a dual reality: one perceived by AI algorithms and another by humans. This method stands in contrast to traditional adversarial attacks that typically focus on modifying pixel values in ways that are either slightly perceptible or entirely imperceptible for both humans and machines.

As developed in the present research, the image algorithm for transparency attacks introduces novel features that exploit the perceptual discrepancies between human vision and machine learning algorithms, particularly in computer vision. At its core, the algorithm harnesses the concept of stealth transparency, where an additional layer - imperceptible to AI but visible to the human eye - is superimposed onto a conventional image. This transparent layer is crafted using iterative image processing techniques that alter the picture in a manner undetectable by standard computer vision models.

When we blend pairs of contrasting image classes, the human and AI perceptible layers can contradict and confuse the model trainer and the end user. We describe the discovered flaw as a transparency attack, owing to the fine-tuning used to set apart the grayscale foreground from the background. The paper investigates the transparency-generating algorithm and catalogs the attack surface both in use cases and for major object detectors (MobileNetV2, YOLOv5), vision language models (GPT4-vision), stable diffusion (Midjourney), and generative adversarial networks (Pix2Pix).

**Architectures and Image Formats.** Previous work [1-8] has examined the role of using transparent layers for digital watermarks, but modern vision algorithms generally regard a flattened image when constructing models. Model builders see the advantage of treating a JPEG the same as a multi-layer RGBA image format like PNG. In addition to ignoring the alpha layer, vision models have gravitated to similar pixel representations in convolutional neural networks (CNNs), transformers, and generative adversarial networks. Broad surveys by Akhtar et al. [1] and Long et al. [2] have examined the landscape of prior vision attacks, setting the stage for further inquiry into new strategies. The emergence of vision transformers, with their inherent robustness characteristics as investigated by Wei et al. [3] and Mahmood et al. [4], has shifted the research focus from conventional CNNs, which are susceptible to such attacks [5]. The development of adversarial attacks that are transferable across various architectures, as evidenced by Ma et al. [6], illustrates the adaptability and potential reach of these attacks. These issues are not merely academic; they carry significant real-world implications, as demonstrated by Zhang et al. [7] in the context of autonomous vehicles and probed by Dujmović et al. [8] from a human vision perspective.

| Attack Image (Background) | Human Visible Alpa Transparent | ChatGPT4-Vision Caption |
|---|---|---|
| 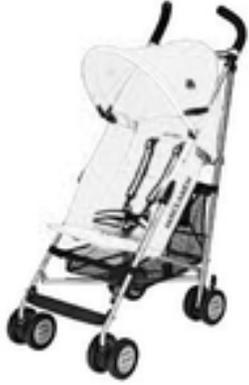 | 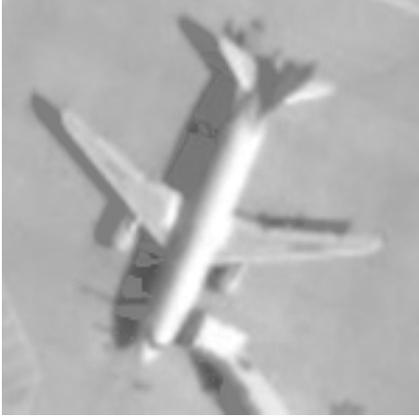 | The image **you've uploaded appears to be that of a baby stroller**, also known as a pushchair or pram. It is a device used to transport an infant or young child from one location to another and is designed to be pushed by a parent or caregiver. This particular stroller seems to be of a compact and lightweight design, which suggests it is easily portable possibly designed for travel or quick trips. |

Table 1. AI vision alignment contradicts the human visible image of an airplane (AI sees a baby carriage)

**Hidden Information in Alpha Layers. Table 1** illustrates the effects of the basic blending algorithm and its distorted caption (right) relative to the middle visible frame that a human sees. The AI correctly reports its view of the background baby stroller (left), which is invisible to human perception (middle). This way, the background image (stroller) can be considered the attack or poisoning source, while the semi-transparent foreground dominates the perceived view (airplane). This broad concept of transparent adversarial examples, explored by Borkar and Chen [9], presents a unique challenge to the established domain of digital watermarking, as initially secured by techniques described by Lin [10], Wolfgang et al. [12], and Singh and Chadha [13]. The traditional safeguards provided by watermarking and steganography, as advanced by Jain and Boaddh [14] and Ansari et al. [15], are now subject to circumvention via such transparent layers. The advent of transparent adversarial attacks raises concerns within sectors reliant on image recognition. For instance, object detection in electro-optical/infrared (EO/IR) and synthetic aperture radar (SAR) imagery, addressed by Lane et al. [16], must now contend with the potential for adversarial concealment. The manipulation of image scaling algorithms, as presented by Xiao et al. [17], underscores the sophistication of these adversarial techniques.

**Potential Consequences of Transparent Vulnerabilities.** As single attack images that might deceive AI detectors, perhaps the most damaging effects center on a long history of hiding information in imagery such as watermarks, steganographic messages and filtering illicit or controversial content. When multiplied across large repositories of training data, the vision systems propagate the distorted labels, and the model output degrades. For instance, if an adversary wants to make a particular target stealthy to satellite views, one can envision poisoning popular large training datasets with a hidden label not visible to the casual model builder. **Table 2** shows a passenger plane in the foreground human view but a mushroom cloud visible to the AI system. When carefully tuned, the blended composite image fools one of the larger vision language models currently deployed to at least 300 million users (ChatGPT-4 Vision). **Table 2** also includes a control that uses the foreground image for the background with no alpha layer for caption verification.

This systematic evaluation of backdoor data poisoning attacks [18], the toxicological impact of data poisoning [19], and the limitations of poisoning to prevent facial recognition [20] have been documented. The generation of invisible data poisoning [21] and subsequent efforts to foster adversarial robustness [22] emphasize the need for new defensive capabilities. Efforts to preclude poisoning attacks using generative models [23], to address attacks in multimodal learning contexts [24], and to characterize such attacks on generalistic AI models [25] indicate a broader recognition of the threat posed by adversarial manipulations. The repercussions of these vulnerabilities extend to applications such as real-time image captioning systems [26], which may generate inaccurate descriptions if the training data is compromised. The integrity of datasets [27] used to train object detection models like YOLOv5 [28] is paramount to the reliability of these systems. The conjunction of language models with vision systems, as exemplified by coupling CLIP with GPT-4 [29], necessitates vigilance to ensure that training and operational data are free from adversarial modifications [30].

**Research Question.** The research hypothesis centers around the effectiveness of this novel algorithm in creating a divergence between what AI 'sees' and what humans perceive in a stacked image pair comprising a background layer and a foreground transparent layer. The hypothesis posits that while the human eye can discern the semi-transparent alpha layer and its contents, AI algorithms, particularly those used in image recognition and captioning, will fail to recognize or misinterpret this layer, focusing solely on the background. *The test involves applying the algorithm to various image pairs and subsequently evaluating the performance of AI models in accurately identifying and describing the contents of these images. The underlying question is whether these AI models can be misled to overlook or mislabel elements present in the transparent layer, thereby testing the algorithm's ability to exploit the inherent limitations of current computer vision technologies.* If validated, this hypothesis could have implications for AI security, digital media, and information dissemination, challenging the existing paradigms of machine perception and its susceptibility to nuanced adversarial tactics. Therefore, the present work aims to extend the resilience of AI systems against adversarial threats that leverage the subtle discrepancies between human and machine vision.

## 2. RESEARCH METHODS

In the attack methodology, we establish a staged image pipeline to evaluate the effectiveness of the transparency attack algorithm across various vision models. The pipeline begins with the careful selection and preparation of grayscale image pairs. Each pair consists of a background image and a foreground image, matched closely in dominant light or dark tones to ensure consistency in the visual theme. This matching is crucial as it forms the basis of the subsequent blending process, where the foreground image, processed to be a transparent layer, is superimposed onto the background image. The transparency level is iteratively adjusted to be imperceptible to AI but discernible to the human eye, thereby setting the stage for the transparency attack. The stages and details of the image layering can be understood in Appendix A: Pseudo-code for the blending algorithm.

| Background Image | Human Visible With Alpha Manipulation | ChatGPT4-Vision Caption |
|---|---|---|
| 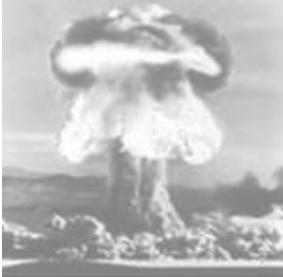 | 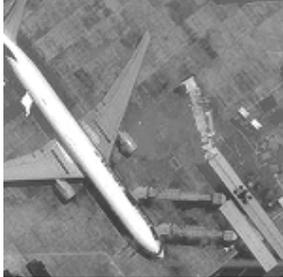 | The image you've uploaded, while blurry, appears to **depict a mushroom cloud**, which is the distinctive cloud of smoke, flame, and debris that results from a large explosion, most notably associated with nuclear detonations. These clouds form as a result of the intense heat causing an updraft that sucks up dust and debris, combined with the condensation of water vapor in the air. Due to the nature of the image, it may represent a historical photograph of a nuclear test or an artistic representation of one. It's important to note the significance of such images and the destructive power they represent. |
| 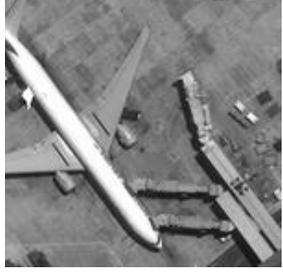 | No alpha layer. | The image you've uploaded appears to be an aerial or satellite **photograph of an airplane** at an airport gate. The perspective is top-down, and you can see the jet bridge connected to the aircraft's door, allowing passengers to board or disembark. The plane's wingspan is prominent, and it's parked on the tarmac, surrounded by ground support equipment and vehicles that service the aircraft between flights. This might include refueling, catering, and baggage handling operations. The image has a grayscale filter, giving it a monochromatic look. |

Table 2. AI vision alignment contradicts the human visible image of an airplane (AI sees an atomic blast)

Once the images are prepared, they are systematically fed into a robust test harness comprising diverse vision algorithms. This test harness includes traditional convolutional neural networks like YOLO, widely used for object detection tasks, and more recent advancements in the field, such as Vision Transformers (ViT) and vision-language models. These models (**Table 3**) have been selected to cover a broad spectrum of AI image analysis capabilities, from object detection and classification to complex captioning tasks. The scale of evaluation is extensive, encompassing a large dataset of image pairs to ensure statistical significance in the results. Each model is tasked with analyzing the blended images, and their outputs are then compared against the known elements of the foreground and background images. The primary objective of this evaluation is to assess the degree to which these algorithms can be deceived by the transparency layer, thereby testing the algorithm's efficacy under diverse AI vision paradigms. This comprehensive approach evaluates the algorithm's ability to exploit the perceptual gaps between human and machine vision.

**2.1 Data Preparation.** The study employed a preprocessing routine to standardize images for subsequent analysis. Each image was converted to grayscale, resized to a uniform dimension of 150x150 pixels, and then transformed into a three-channel RGB format. This preprocessing ensured consistency across images, facilitating the comparison of features and the application of uniform transformations in subsequent steps.

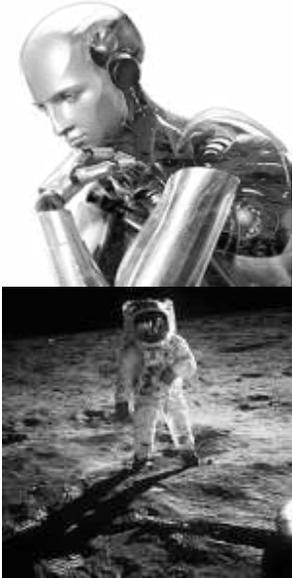

| Human Visible/Attack Image | Image-to-Text Models Tested for Successful Attack as Recognizing Only the Apollo 11 Hidden Image |
|---|---|
| | ChatGPT4-Vision (27JAN2024) |
| | Midjourney /describe version 6 (stable diffusion) |
| | Ultralytics/YOLOv5 |
| | Salesforce/blip-image-captioning-large |
| | microsoft/git-large |
| | Abdou/vit-swin-base-224-gpt2-image-captioning |
| | nnpy/blip-image-captioning |
| | microsoft/resnet-50 |
| | timm/mobilenetv3_large_100.ra_in1k |
| | nvidia/mit-b0 |
| | timm/inception_v3.gluon_in1k |
| | facebook/deit-base-patch16-224 |
| | timbrooks/instruct-pix2pix |
| | microsoft/resnet-18 |
| | facebook/convnext-large-224 |
| | apple/mobilevitv2-1.0-imagenet1k-256 |

**Table 3. Models Tested as Susceptible to Hidden Transparency Attacks Include Foundational Object Classifiers, Vision Language Models, and Diffusion Image-to-Image Models**

**2.2 Blending Algorithm.** A custom image blending algorithm was developed to merge a target image with a predefined background image (**Appendix A**). The background image served as a baseline for transparency manipulation. The algorithm implemented an iterative optimization process that adjusted the alpha transparency layer to minimize the mean squared error (MSE) between the blended and target images. The optimization used the Adam optimizer, a stochastic gradient descent method known for efficiently handling sparse gradients on noisy problems.

The blending process was as follows:

1. An initial alpha layer was created, filled with ones, indicating full opacity across the image.
2. A white background was generated as a canvas for the blending process.
3. For a fixed number of iterations, the algorithm performed the following steps:
   a. Blended the background and white images using the alpha layer to determine transparency.
   b. Computed the MSE loss between the blended image and the target image.
   c. Backpropagated the error to update the alpha layer through gradient descent.
   d. Periodically logged the progress, detailing the step number and current loss value.

After completing the iterations, the algorithm produces a final image that, to the human eye, closely resembles the target image, all while preserving the original background. This blending achieves a dual perception: visually, it appears as the target image to humans, yet computer vision algorithms identify it as the background image.

**2.3 Post-Processing and Scoring Misclassifications.** The resulting blended images were saved in PNG format with an appended tag to the original filename. We submit the doctored images to object classifiers like traditional YOLOv5 and augmented vision language models (ViT Transformers, Generative Pretrained Transformers GPT-Vision), as shown in **Table 3**.

## 3. RESULTS

The main results of our approach demonstrate the breadth of the vision processing vulnerabilities, both in diverse algorithms and across wide-ranging fields of application. The single-image version allows the transmission of hidden information and bypasses standard methods of content filters or human examination. The large image repositories allow for classic data poisoning scenarios where an entire image class, like a military tank, could be masked from satellite detection. **Table 4** highlights some of the initial explorations of the transparency attack surface. In these cases, the attack image is regarded as the first ever sent over the internet based on a cabaret theater show featured at the CERN laboratory in the 1990s.

| Application | Image Pairs (Target Foreground + Background) | Caption and Mislabeling Source |
|---|---|---|
| **Dataset Poisoning. Mislabeling Noise.** | 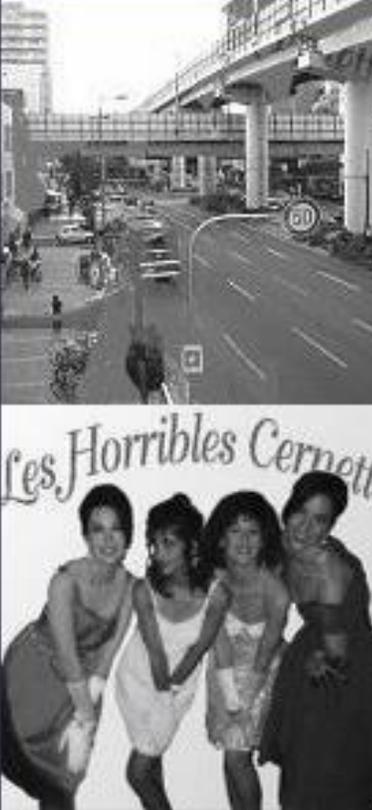 | This image appears to be a black-and-white photograph of **four women posing together.** They are dressed in costumes that suggest a performance or band theme, as indicated by the text *"Les Horribles Cernettes"* on the top of the image. The quality of the photograph seems dated, which could imply that it's from an earlier time, possibly the late 20th century. The low resolution and high compression artifacts suggest that this might be an image that has been copied or shared multiple times, typical of early digital photographs. The women are smiling and appear to be enjoying themselves, contributing to a lively and cheerful atmosphere. *(ChatGPT-4/ Vision, 25JAN2024 version)* |

| Application | Image Pairs (Target Foreground + Background) | Caption and Mislabeling Source |
|---|---|---|
| **Hidden Message, Alpha Layer Steganography**<br><br>The person in the foreground seems to be casting a fishing line, as indicated by the stance and the arm position, which is typical of someone who is fishing. The second, smaller figure in the background also appears to be doing the same. The surrounding lines suggest a tranquil outdoor setting, possibly near a lake or river. This scene captures a common leisure activity that is both relaxing and focused, conveying a sense of calm and patience that fishing typically requires. | 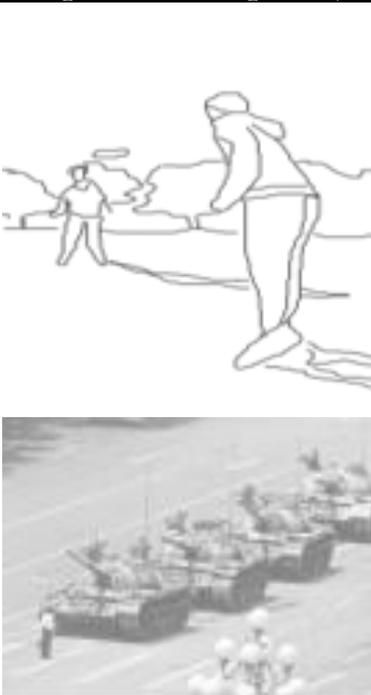 | The image you've referred to depicts the iconic and historic moment known **as the "Tank Man" incident**. It occurred during the Chinese government's crackdown on the pro-democracy protests in Tiananmen Square in Beijing on June 5, 1989. The photo shows a single man, carrying shopping bags, standing in front of a column of tanks, preventing them from moving forward. The "Tank Man" has become a symbol of resistance to oppression worldwide, and the image is widely recognized as a powerful act of peaceful defiance. The identity of the man and his fate after the event remain unknown. The image is a testament to the impact an individual can have against a powerful force.<br>*(ChatGPT-4/ Vision, 25JAN2024 version)* |
| **Fooling Aerial Classifier; Mistake Baseball Field for a Mushroom Cloud or Volcano** | 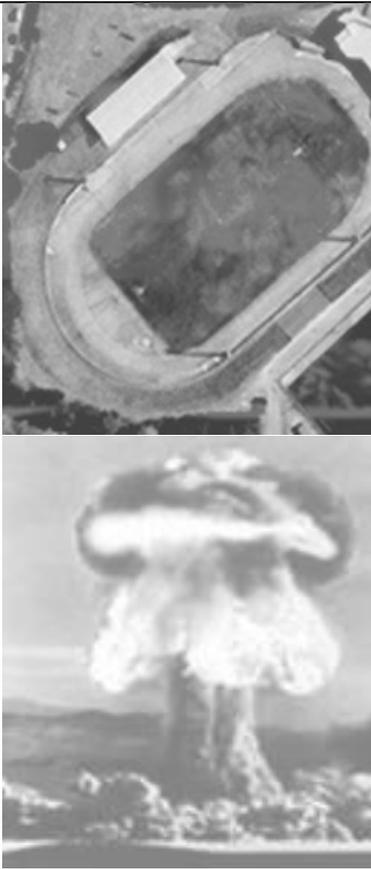 | Class, Confidence<br>volcano 0.06<br>dromedary 0.05<br>frilled-necked lizard 0.03<br>seashore 0.02<br>mushroom 0.02<br><br>(YOLOv5 Classifier) |

| Application | Image Pairs (Target Foreground + Background) | Caption and Mislabeling Source |
|---|---|---|
| **Fooling Facial Recognition and Identification** | 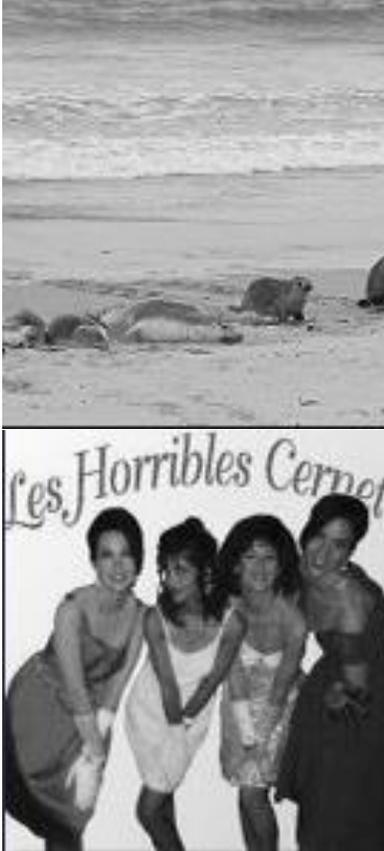 | The image appears to be a black and white photograph of four women posing together. The women are dressed in what looks like evening wear, with sleeveless dresses and hairstyles that suggest the photograph may be from a few decades ago. They are smiling and seem to be enjoying themselves. Behind them, the text "Les Horribles Cernettes" is visible, suggesting that they may be part of a group or band by that name. The quality of the image is somewhat grainy which is characteristic of older photographs or those that have been digitized with low resolution. *(ChatGPT-4/ Vision, 25JAN2024 version)* |
| **Sentiment and Emotion Mis- categorization**<br><br>what is the mood (sad, joy, surprise) and sentiment (positive, negative) of this image?<br><br>*(ChatGPT-4/ Vision, 25JAN2024 version)* | 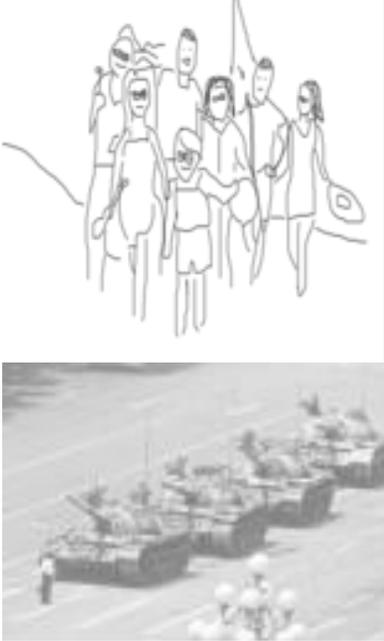 | The image is quite blurry, but it appears to show a military parade or procession with tanks and soldiers, typically associated with a display of strength and national pride. Such events are usually intended to evoke feelings of patriotism and confidence in the country's military capabilities, so the intended mood might be solemn or prideful rather than joyful or sad.<br><br>As for sentiment, in the context of a military parade, it is usually intended to be positive from the perspective of the country showcasing its military. However, the sentiment could be perceived differently by others, depending on their personal views, context, and the current political climate. Without clearer details or more context, it's challenging to determine a precise mood or sentiment. |

| Application | Image Pairs (Target Foreground + Background) | Caption and Mislabeling Source |
|---|---|---|
| Object recognition deception | 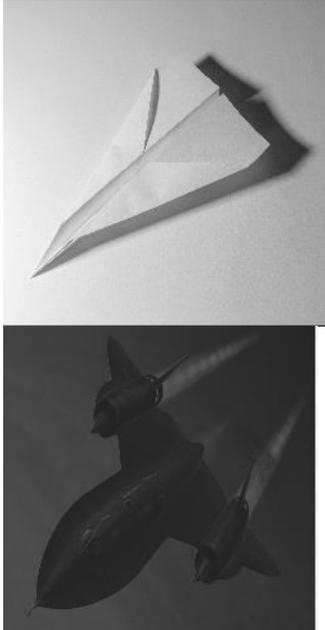 | The image shows a black and white photo of a vehicle that resembles an advanced aircraft or spaceship. It features a sleek, streamlined body with a pointed nose, two wing-like structures on the sides and what appears to be engines at the rear, possibly emitting trails. The design is futuristic, and the aircraft is capture in flights, angled slightly upwards as fi ascending or in high-speed travel. The lighting and shading five it a dramatic, perhaps stealthy appearance. *(ChatGPT-4/ Vision, 25JAN2024 version)* |

**Table 4. Example Application Demonstrations for Transparency Attack with Image Pairs and Mislabeled Outputs**

### 4. DISCUSSION

The research has demonstrated a novel computer vision vulnerability where imperceptible layers—stealth transparency—cause misinterpretation by AI systems while remaining undetectable to human perception. This approach has several potential applications and consequences. The consequences of these applications are far-reaching and underscore the need for robust security measures and monitoring of AI systems to ensure they interpret visual data as intended. Furthermore, this work highlights the importance of designing AI systems that can detect and understand context as adeptly as humans, ensuring that such attacks do not easily deceive them.

For concreteness, **Figure 2** presents a malicious exploit where a brand name (BMW) watermarks a competitor's image (Ferrari) or vice versa. One can extend this polar example to a host of areas where a motivated attacker may identify some notoriety, financial gain, or defacement value in altering a well-known or widely distributed cover image with an invisible background only seen or cataloged by machines. In this way, one can imagine a digital search engine optimization for images, where the indexing algorithm and rank may motivate the attacker to display a deceptive front and contradictory back view. Another recognizable illustration might be mislabeling all "Coke" images with a "Pepsi" layer seen only by the search engine indexer. Thus, when the unknowing keyword searcher asks for representative brand images for "Coke," the competitor's image appears mismatched to the human eye ("Pepsi") but well-matched to the image search engine ("Coke") algorithm. Below, we outline some of the broad categories of affected computer vision systems presently in development or already deployed with this suspected blind spot or vulnerability.

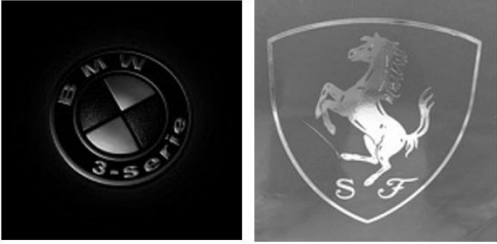

*Figure 2. Example Background Images for Logo Transparency Attacks (BMW vs. Ferrari)*

1. **Security Systems**:
    - **Facial Recognition Evasion**: Attackers could employ stealth transparency to images to evade facial recognition systems without noticeable alterations to the human observer.
    - **Surveillance Avoidance**: Imperceptible layers could be used to prevent surveillance systems from correctly identifying or tracking individuals or objects.
2. **Digital Media and Watermarking**:
    - **Digital Rights Management (DRM)**: Stealth transparency could bypass DRM systems relying on computer vision to detect watermarked media.
    - **Anti-Piracy Efforts**: The effectiveness of anti-piracy watermarking could be compromised, allowing pirated media to avoid detection.
3. **Content Filtering and Moderation**:
    - **Censorship Circumvention**: Users might use imperceptible layers to share content that automatic moderation systems would otherwise flag.
    - **Misinformation Spread**: It could be used to disseminate misinformation by slightly altering images that lead AI systems to generate misleading captions or descriptions.
4. **Machine Learning and Data Science**:
    - **Dataset Poisoning**: As demonstrated in the paper, attackers could poison training datasets, leading to inaccurately trained models or models that fail to generalize well.
    - **Model Robustness Testing**: This method could test the robustness of computer vision models against adversarial attacks.
5. **Automotive and Drone Technology**:
    - **Autonomous Vehicle Misdirection**: Imperceptible layers could be used to create signs or signals that mislead autonomous vehicles while appearing normal to human drivers.
    - **Drone Vision Disruption**: Drones relying on computer vision for navigation could be confused by imperceptible layers that disrupt their understanding of the environment.
6. **E-Commerce and Retail**:
    - **Product Misclassification**: Online retail systems that rely on image recognition for product listing could be manipulated to misclassify items.
    - **Augmented Reality (AR) Shopping Experiences**: AR systems in retail settings might be tricked into displaying incorrect product information due to imperceptible alterations.
7. **Healthcare**:
    - **Medical Imaging Analysis**: Imperceptible layers could be used maliciously to alter medical images so that AI diagnostics would misinterpret them, with potentially dangerous consequences.
8. **Legal and Forensic Analysis**:
    - **Evidence Tampering**: Imperceptible layers could subtly alter images used as legal evidence, potentially misleading investigations.
    - **Forensic Authenticity Checks**: The reliability of forensic tools that analyze images for authenticity or tampering could be compromised.
9. **Art and Design**:
    - **Artistic Expression**: Artists might use imperceptible layers as a form of expression, creating works that are perceived differently by AI and human observers.
    - **Intellectual Property Theft**: Using stealth transparency, designers could hide their work from being stolen by AI-driven design scraping tools.

These applications, while innovative, also raise ethical and security concerns. They could be used maliciously, such as embedding undetectable propaganda in images or creating undetectable fraudulent documents. Therefore, alongside the development of such techniques, there must be parallel advancements in detection capabilities to maintain a balance between the beneficial uses of technology and the protection against its misuse. The described vulnerability has significant implications for steganography and image source tracking. **Table 5** shows multiple example logo attacks with hidden images and deceptive advertising or brand infringements. We demonstrate the vulnerable AI vision model in the first column, the composite or blended image in the center, and an attack image in the background of the Ferrari logo (**Figure 2**). The conclusion of the test matrix captures a mislabeled or deceitful image caption, which marks all the candidates as effectively "spam" advertising for the Ferrari horse logo.

| AI Model (huggingface.co) | Human Visible Image | AI Caption |
|---|---|---|
| Salesforce/blip-image-captioning-large | 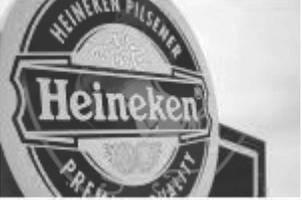 | a close up of a black and white photo of a ferrari logo |
| Salesforce/blip-image-captioning-large | 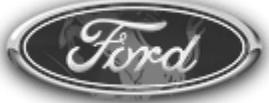 | a close up of a black and white photo of a ferrari logo |
| ChatGPT4-Vision (27JAN2024) | 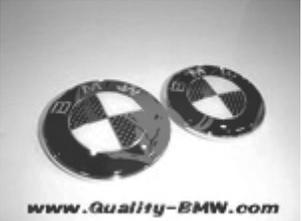 | The image you've uploaded is the logo of Ferrari, an Italian luxury sports car manufacturer. The prancing horse is the symbol of Ferrari, and the initials "S F" stand for Scuderia Ferrari, which is the racing division of the luxury Italian auto manufacturer. The logo is typically associated with a yellow background, and sometimes includes the Italian national colors—green, white, and red—above the horse. |
| Midjourney v.6 | 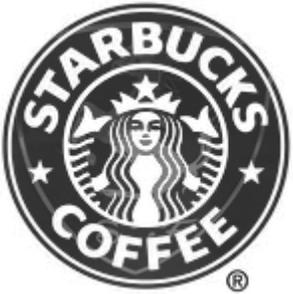 | the s ferrari logo on a black, in the style of frank frazetta, dariusz klimczak, ryohei fuke, dark silver, etching/engraving, cluj school, group f/64<br><br>ferrari logo on black in size, in the style of george stubbs, frank frazetta, grisaille, shigenori soejima, toonami, layered translucency, group f/64<br><br>the logo of ferrari on a black bag, in the style of farm security administration aesthetics, mythical beasts, sepia tone, somber mood, translucent color, realistic equine paintings, group f/64 |

| AI Model (huggingface.co) | Human Visible Image | AI Caption |
|---|---|---|
| | | ferrari and company logo on a black canvas bag, in the style of sepia tone, farm security administration aesthetics, somber mood, animated gifs, silver, sharp focus, group f/64 |
| "Image to Video" Creates Logo Blend<br><br>https://app.runwayml.com/<br><br>Creates a 4 second video extended from the miscued background of Ferrari logo | 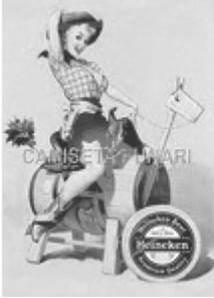 | 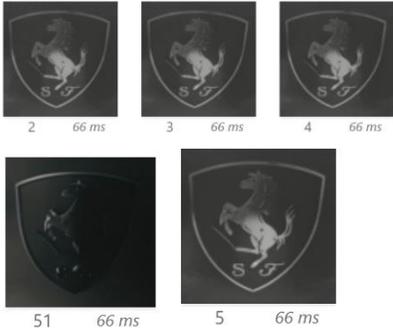 |
| YOLOv5 Classifier<br><br>https://github.com/ultralytics/ultralytics | 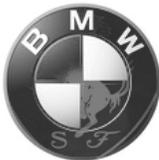 | 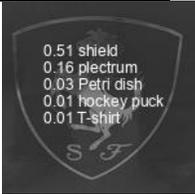 |
| ViLMedic/captioning_baseline | 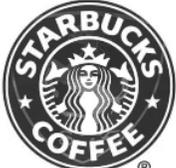 | a painting of a cow with a cross on it |
| microsoft/git-base-coco | 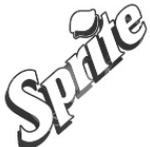 | a horse emblem on the back of a car in black and white. |
| nlpconnect/vit-gpt2-image-captioning | 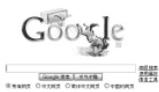 | a painting of a cow with a cross on it |
| Zayn/AICVTG_What_if_a_machine_could_create_captions_automatically | 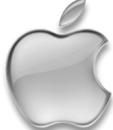 | a painting of a cow with a cross on it |
| Sof22/image-caption-large-copy | 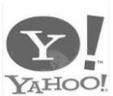 | a close up of a black and white photo of a ferrari logo on a black background with a white horse |
| nnpy/blip-image-captioning | 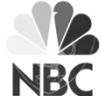 | output : a black and white picture of a horse with the letters s. f on it. |

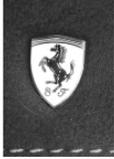

| AI Model (huggingface.co) | Human Visible Image | AI Caption |
|---|---|---|
| ChatGPT4-Vision (27JAN2024)<br><br>Reverse instance where background is BMW and foreground is Ferrari | | The image you've uploaded shows the logo of BMW, specifically noting the "3-serie," which refers to the BMW 3 Series. The BMW 3 Series is a line of compact executive cars manufactured by the German automaker BMW, and the logo features the distinctive BMW colors and emblem, which is often erroneously thought to represent a spinning propeller. The design actually originates from the Bavarian state flag, which is blue and white, reflecting the company's origins in Bavaria, Germany. The BMW 3 Series itself is known for its performance and handling, and it is one of the company's most popular models. |

**Table 5. Example, Logo Attacks with Hidden Images and Deceptive Advertising or Brand Infringements**

As illustrated in **Figure 3**, the transparency attack extends beyond single images to the underlying detection and captioning models when a repository is poisoned and integrated into future training cycles. We trained a MobileNetV2 classifier with identical input images of passenger airplanes, except the second class, "Blended Attacks," hides an underlying mushroom cloud. The resulting MobileNetV2 model can mistakenly distinguish between identical image sets with 100% accuracy. The novel effect is the underlying CNN sees the blended or attack class as distinct and can recognize the background layer despite its invisibility to the model trainer. This scenario propagates the vulnerability beyond single "pranks" into a broader and persistent distortion of the trained model for object detection. The attacker can hide an entire class of labeled images or make an offensive image into a benign one to a content filtering system. The surreptitious infusion of doctored images into large datasets carries profound implications for training neural networks, as evidenced by the poisoning of datasets aimed at compromising the integrity of models like MobileNetV2. An attacker can ingeniously manipulate the model's learning process by embedding an imperceptible layer into the images. Despite their invisibility to the human eye and the model trainer, the underlying convolutional neural network (CNN) can accurately distinguish these altered images. This phenomenon reveals a vulnerability that transcends isolated mischief, seeding a systemic distortion within the model's object detection capabilities. Such an attack could subversively conceal an entire category of images within a dataset or masquerade offensive content as innocuous, effectively duping content filtering systems. The implication of this is a silent yet potent capability to disrupt AI-driven moderation and surveillance, introducing a novel class of threats that could persistently undermine the veracity of automated systems and the broader digital ecosystem they support.

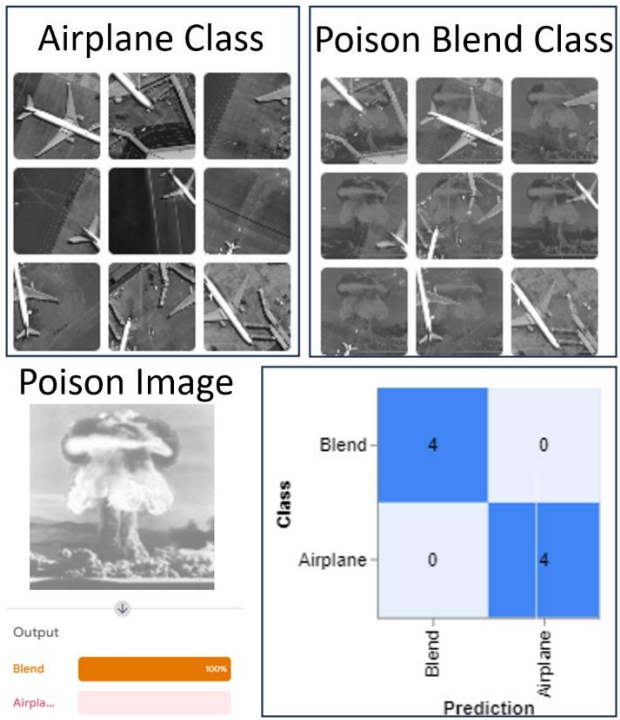

*Figure 3. MobileNetV2 Classifier Poisoned by Hidden Attack Image of Mushroom Cloud.*

Beyond leaving a persistent corrupt dataset and model, the forward propagation of the model in future predictions or transfer learning enhances the vulnerability's effects. The prospect of using imperceptible image layers to deceive generative adversarial networks (GANs), such as Pix2Pix, introduces a fascinating and somewhat alarming dynamic in AI and machine learning.

| Image Visible | Attack Image | Instruct-Pix2Pix |
|---|---|---|
| 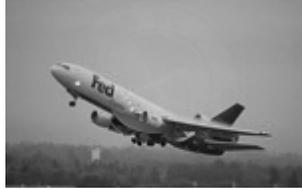 | 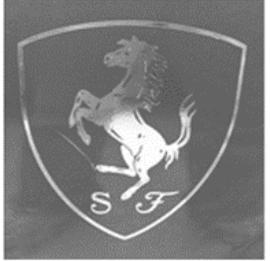 | 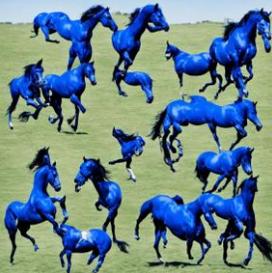 |
| https://huggingface.co/timbrooks/instruct-pix2pix **Human sees this image** | **AI sees this hidden image and modifies it** | *"Make the horse blue"* |

Table 6. Attack image poisons Image-to-Image inference with instructions and Propagating and Persistent Mislabeling

In such a scenario, an image is doctored with a layer invisible to the human eye but detectable by the AI. When this AI, trained to modify pictures based on its learning, receives the command to alter the doctored image, it does not work on the image as perceived by human observers. Instead, it interacts with the hidden layer, applying transformations to an unseen aspect of the image. This discrepancy creates a dichotomy between the human and machine perception of the same image, leading to results that can be unexpected or entirely misleading from a human perspective.

To extend this forward propagation beyond object classifiers, we tested a simple, instructible image generator based on the Pix2Pix architecture **(Table 6).** As a promising model for visual style transfer, we submitted to a trained Pix2Pix GAN the blended image of a FedEx passenger plane on top of a background hidden image of the Ferrari horse logo in Figure 2. We prompt the image generator to make the horse blue, which is non-sensical instruction to the human viewer of the airplane, but the model succeeds in modifying what it sees as the whole input image, which is the Ferrari horse logo. This manipulation of the Pix2Pix GAN exploits the nuanced ways these networks learn and interpret image data, revealing a potential exploit in how GANs are trained and operate. Since GANs typically learn to generate and modify images based on the data they have been fed, an imperceptible layer effectively trains the network to recognize and process images differently than a human would. This propagating attack could have far-reaching consequences, from creating a hidden communication channel that only AI can interpret to undermining the reliability of AI systems in critical applications, such as medical imaging analysis, where an unseen layer might cause a misdiagnosis, or in autonomous vehicles, where it could lead to incorrect scene interpretation. The ramifications of such a technique underscore the importance of robustness in AI training and the need for safeguards against such subversive tactics.

## 5. CONCLUSION

Future work should extend the attack surface and harden the algorithm to detect the background and foreground matches in a tailored blend operation. Since the attack relies on grayscale layers, further effort might be needed to extend the image pairing into other PNG formats or video color layers. The major limitation of the transparency attack is the low success rate when the human viewer's background theme is not light by default or at least a close match to the transparent foreground and hidden background layers. When mismatched, the background becomes visible to the human eye and the vision algorithm.

As described here, the application of an imperceptible image layer in RGBA formats such as PNG presents a novel frontier for steganography and image source tracking. Such a technique offers a discreet yet robust method for embedding data within an image, facilitating new steganographic methods that elude detection by machine learning models while remaining accessible to informed human parties. It serves as a clandestine channel for secure communication, particularly valuable in heavily monitored environments, allowing for transmitting sensitive information without the risk of interception by AI systems. Moreover, it redefines digital watermarking by enabling the concealment of ownership marks from both unauthorized AI detection and the human eye, preserving the visual integrity of digital media. The imperceptible layer becomes a silent guardian of authenticity, providing a means for invisible tracing and anti-theft measures, empowering photographers and digital artists with a non-intrusive claim to their work. In the critical arenas of forensics and law enforcement, it allows for embedding traceable information without compromising the evidentiary value of images, enhancing the tracking of content distribution while ensuring image authenticity. This innovative approach promises to bolster the security and integrity of digital images, fostering advancements in secure communication and digital rights management.

In conclusion, this study has developed a novel algorithm that leverages stealth transparency to create a divergence in perception between human observers and AI-driven vision systems. The algorithm manipulates grayscale images by superimposing a transparent layer, which, while perceptible to the human eye, remains largely undetected or misinterpreted by a range of AI vision models. The algorithm was rigorously tested across various AI systems, including traditional convolutional networks and advanced vision transformers, through a comprehensive and robust experimental pipeline. The results were striking: the AI models consistently failed to recognize or accurately classify elements within the transparent layer, focusing instead on the background image. This significant finding validates the initial hypothesis and underscores the potential vulnerabilities in current AI vision technologies. The implications of these results are far-reaching, opening avenues for further research in AI security and the development of more resilient machine vision algorithms. Additionally, the study highlights the importance of considering human-like perceptual capabilities in AI systems to bridge the gap in visual interpretation between humans and machines. The novel image manipulation technique maintains visual integrity for human observers while creating a parallel and detectable reality for AI, diverging from conventional adversarial approaches by exclusively targeting machine perception.

### ACKNOWLEDGMENTS
The authors thank the PeopleTec Technical Fellows program for encouragement and project assistance.

## Appendix A: Pseudo-Code for Novel Image Blending Algorithm

```
Procedure LoadAndPreprocessImage(path, size):
    Read image from path as grayscale
    Resize image to given size
    Convert image color from grayscale to RGB
    Return image

Procedure BlendImages(target_images_dir, background_path, size, steps, learning_rate):
    background_image <- LoadAndPreprocessImage(background_path, size)
    background_tensor <- ConvertToTensor(background_image) * 0.5

    For each filename in target_images_dir:
        If filename ends with ".jpg":
            target_image <- LoadAndPreprocessImage(PathJoin(target_images_dir, filename), size)
            target_tensor <- ConvertToTensor(target_image)
            alpha <- InitializeTensorWithOnes(DimensionOf(background_tensor))

            optimizer <- InitializeAdamOptimizer(alpha, learning_rate)

            white_background <- InitializeTensorWithOnesLike(background_tensor)

            For step from 0 to steps:
                ResetGradients(optimizer)

                blended_image <- alpha * background_tensor + (1 - alpha) * white_background

                loss <- ComputeMSELoss(blended_image, target_tensor)
                BackpropagateLoss(loss)

                UpdateOptimizer(optimizer)

                If step modulo 100 equals 0:
                    LogStepProgress(filename, step, loss)

            SaveBlendedImage(background_tensor, alpha, target_images_dir, filename)

    Log("Processing complete.")

Main:
    Define target_images_dir, background_path, size, steps, learning_rate
    BlendImages(target_images_dir, background_path, size, steps, learning_rate)
```